%% file: root.tex
\documentclass[letterpaper, 10 pt, journal, twoside]{IEEEtran}

\input{setup}

\begin{document}
    

\author{Tim Pfeifer, Sven Lange and Peter Protzel%
	\thanks{Manuscript received: August 25, 2020; Revised December 1, 2020; Accepted March 1, 2021.
            This paper was recommended for publication by Editor Allison Okamura upon evaluation of the Associate Editor and Reviewers' comments.
		    This work was supported by "Bundesministerium f\"ur Wirtschaft und Energie" (German Federal Ministry for Economic Affairs and Energy).}
	\thanks{The authors are with the Faculty of Electrical Engineering and Information Technology,
            Technische Universität Chemnitz, Germany,
            (e-mail: tim.pfeifer@etit.tu-chemnitz.de; sven.lange@etit.tu-chemnitz.de; peter.protzel@etit.tu-chemnitz.de)}
	\thanks{Digital Object Identifier (DOI): \href{https://dx.doi.org/10.1109/LRA.2021.3067307}{10.1109/LRA.2021.3067307}}
}

\title{Advancing Mixture Models for Least Squares Optimization}
\maketitle

\fancyfoot{}
\fancyhead[OL]{ 
    \footnotesize
    Published in IEEE Robotics and Automation Letters (RA-L), 2021. ACCEPTED VERSION\\
    \tiny
    \copyright 2021 IEEE. Personal use of this material is permitted.  Permission from IEEE must be obtained for all other uses, in any current or future media, including reprinting/republishing this material for advertising or promotional purposes, creating new collective works, for resale or redistribution to servers or lists, or reuse of any copyrighted component of this work in other works.
}
\addtolength{\headheight}{\baselineskip}
\thispagestyle{fancy}
\pagestyle{empty}

\input{inc/abstract}


\input{inc/intro.tex}

\input{inc/prior.tex}

\input{inc/gmm.tex}

\input{inc/eval.tex}

\input{inc/discussion.tex}
 
\input{inc/conclusion.tex}

\input{inc/appendix.tex}

\bibliographystyle{IEEEtran}
\bibliography{IEEEabrv,tipf,lasve}

\end{document}

%% file: setup.tex
\usepackage[utf8]{inputenc}
\usepackage[T1]{fontenc}

\usepackage{microtype}

\usepackage{csquotes}

\usepackage[pdftex]{graphicx}

\usepackage{amsmath} 
\usepackage{amssymb} 
\usepackage{siunitx}

\usepackage{booktabs}
\usepackage{multirow}
\usepackage{multicol}
\usepackage[table]{xcolor} 

\usepackage{fancyhdr}

\usepackage{hyperref}
\hypersetup{
    pdftitle={Advancing Mixture Models for Least Squares Optimization},
    pdfauthor={Tim Pfeifer, Sven Lange and Peter Protzel},
    pdfsubject={A novel Gaussian Mixture representation for nonlinear least squares},
    pdfkeywords={Sensor Fusion, Optimization, Gaussian Mixtures, Least Squares, State Estimation, Factor Graphs}
}

\def\equationautorefname~#1\null{(#1)\null}

\input{inc/mathstuff}

\input{inc/mathstuff_tipf}

%% file: inc/mathstuff.tex
\newcommand{\vect}[1]{\mathbf{ #1}}

\newcommand{\vectg}[1]{{\boldsymbol{ #1}}}

\newcommand{\T}{^\mathsf{T}}

\newcommand{\argmin}{\operatornamewithlimits{argmin}}

\newcommand{\argmax}{\operatornamewithlimits{argmax}}

\newcommand{\normal}[2]{\mathcal{N}\left(#1\, |\, #2\right)}

\newcommand{\uniform}[2]{\mathcal{U}\left(#1\, |\, #2\right)}

\newcommand{\pfrac}[2]{\frac{\partial #1}{\partial #2}}  

\newcommand{\smd}[2]{\left\| #1 \right\|^2_{#2}}

\newcommand{\vR}{\vect{R}}

\newcommand{\vm}{\vect{m}}

\newcommand{\vr}{\vect{r}}

\newcommand{\vx}{\vect{x}}

\newcommand{\vmu}{\vectg{\mu}}

\newcommand{\vSigma}{\vectg{\Sigma}}

\newcommand{\cF}{\mathcal{F}}

\newcommand{\cM}{\mathcal{M}}

\DeclareMathAlphabet\mathbfcal{OMS}{cmsy}{b}{n}


%% file: inc/mathstuff_tipf.tex
\makeatletter
\renewcommand*\env@matrix[1][\arraystretch]{%
	\edef\arraystretch{#1}%
	\hskip -\arraycolsep
	\let\@ifnextchar\new@ifnextchar
	\array{*\c@MaxMatrixCols c}}
\makeatother

\newcommand{\mat}[1]{{\vect{\MakeUppercase{#1}}}}
\newcommand{\set}[1]{\mathcal{{\MakeUppercase{#1}}}}

\newcommand{\est}[1]{\hat{#1}}

\newcommand{\given}{\middle|}

\newcommand{\exponent}{e}
\newcommand{\scaling}{s}

\DeclareMathOperator{\lse}{sLSE}

\DeclareMathOperator{\zero}{\boldsymbol{0}}
\DeclareMathOperator{\half}{\frac{1}{2}}

\newcommand{\iGMM}{l}
\newcommand{\iMax}{k}
\newcommand{\iRes}{n}

\newcommand{\iGMMEnd}{{\MakeUppercase{\iGMM}}}

\newcommand{\iResEnd}{{\MakeUppercase{\iRes}}}

\newcommand{\sumGMM}{\sum_{\iGMM=1}^{\iGMMEnd}}

\newcommand{\maxGMM}{\max_{\iGMM}}
\newcommand{\argmaxGMM}{\argmax_{\iGMM}}

\newcommand{\prob}[1]{P\left( #1 \right)}
\newcommand{\State}{\set{x}}
\newcommand{\state}{\vect{x}}
\newcommand{\Measurement}{\set{z}}
\newcommand{\measurement}{\vect{z}}

\newcommand{\StateEst}{\set{\est{x}}}

\newcommand{\stateEst}{\vect{\est{x}}}

\newcommand{\residual}{\vect{r}}

\newcommand{\loss}{\boldsymbol{\rho}}
\newcommand{\lossMM}{\boldsymbol{\rho}_{m}}
\newcommand{\lossSM}{\rho_{s}}
\newcommand{\lossMSM}{\boldsymbol{\rho}_{ms}}

\newcommand{\normMM}{\gamma_{m}}
\newcommand{\normSM}{\gamma_{s}}
\newcommand{\normMSM}{\gamma_{ms}}

\newcommand{\Normal}[1]{\mathcal{N}\left( #1 \right)}
\newcommand{\mean}{\boldsymbol{\mu}}
\newcommand{\cov}{\boldsymbol{\Sigma}}
\newcommand{\covEst}{\boldsymbol{\est{\Sigma}}}
\newcommand{\info}{\boldsymbol{\mathcal{I}}}
\newcommand{\sqrtinfo}{\info^{\frac{1}{2}}}

\newcommand{\weight}{w}

%% file: inc/abstract.tex
\begin{abstract}
	Gaussian mixtures are a powerful and widely used tool to model non-Gaussian estimation problems.
	They are able to describe measurement errors that follow arbitrary distributions and can represent ambiguity in assignment tasks like point set registration or tracking. 
	However, using them with common least squares solvers is still difficult.
	Existing approaches are either approximations of the true mixture or prone to convergence issues due to their strong nonlinearity.
	We propose a novel least squares representation of a Gaussian mixture, which is an exact and almost linear model of the corresponding log-likelihood.
	Our approach provides an efficient, accurate and flexible model for many probabilistic estimation problems and can be used as cost function for least squares solvers. 
	We demonstrate its superior performance in various Monte Carlo experiments, including different kinds of point set registration.
	Our implementation is available as open source code for the state-of-the-art solvers Ceres and GTSAM.
\end{abstract}

%% file: inc/intro.tex

\section{Introduction}

\IEEEPARstart{W}{ould}
it not be great to have one probabilistic tool to describe arbitrary distributions that arise from simultaneous localization and mapping (SLAM), tracking or even point set registration?
We show how to include Gaussian mixture models into the typical nonlinear least squares solvers that are used for these problems today.

State estimation in robotics is based on deterministic models, that describe the physical relation between measurements and states.
Due to the absence of perfect measurements, probabilistic models are required additionally to describe the related uncertainty.
For these models, the Gaussian distribution is the first choice, since it allows efficient least squares optimization.

However, sensor information is often corrupted by outliers that do not follow a Gaussian: Vision can blur, odometry can slip and GNSS can be affected by reflections and multipath.
Even if the sensor noise itself is Gaussian, ambiguity can lead to multimodal densities for SLAM or point set registration problems.
Without better probabilistic models, state estimation is likely to fail in these cases. 

One of the most popular models for robust estimation is the Gaussian mixture model (GMM), which is composed of a sum of multiple weighted Gaussian components.
In difference to the also common M-estimators, GMMs are preferred when outliers follow a systematic pattern or ambiguity has to be represented.
They are flexible enough to represent asymmetry, heavy tails and skewness, but close enough to a single Gaussian to allow sufficient convergence using nonlinear least squares.

There already exist algorithms to include GMMs in nonlinear least squares optimization, like Max-Mixture (MM) \cite{Olson2013} or the approach proposed in \cite{Rosen2013}, in this work referred as Sum-Mixture (SM).
However, both of them come with individual drawbacks in terms of model accuracy or convergence as we explain in \autoref{sec:prior}.

\begin{figure}[t]
    \centering
    \includegraphics[width=0.975\linewidth]{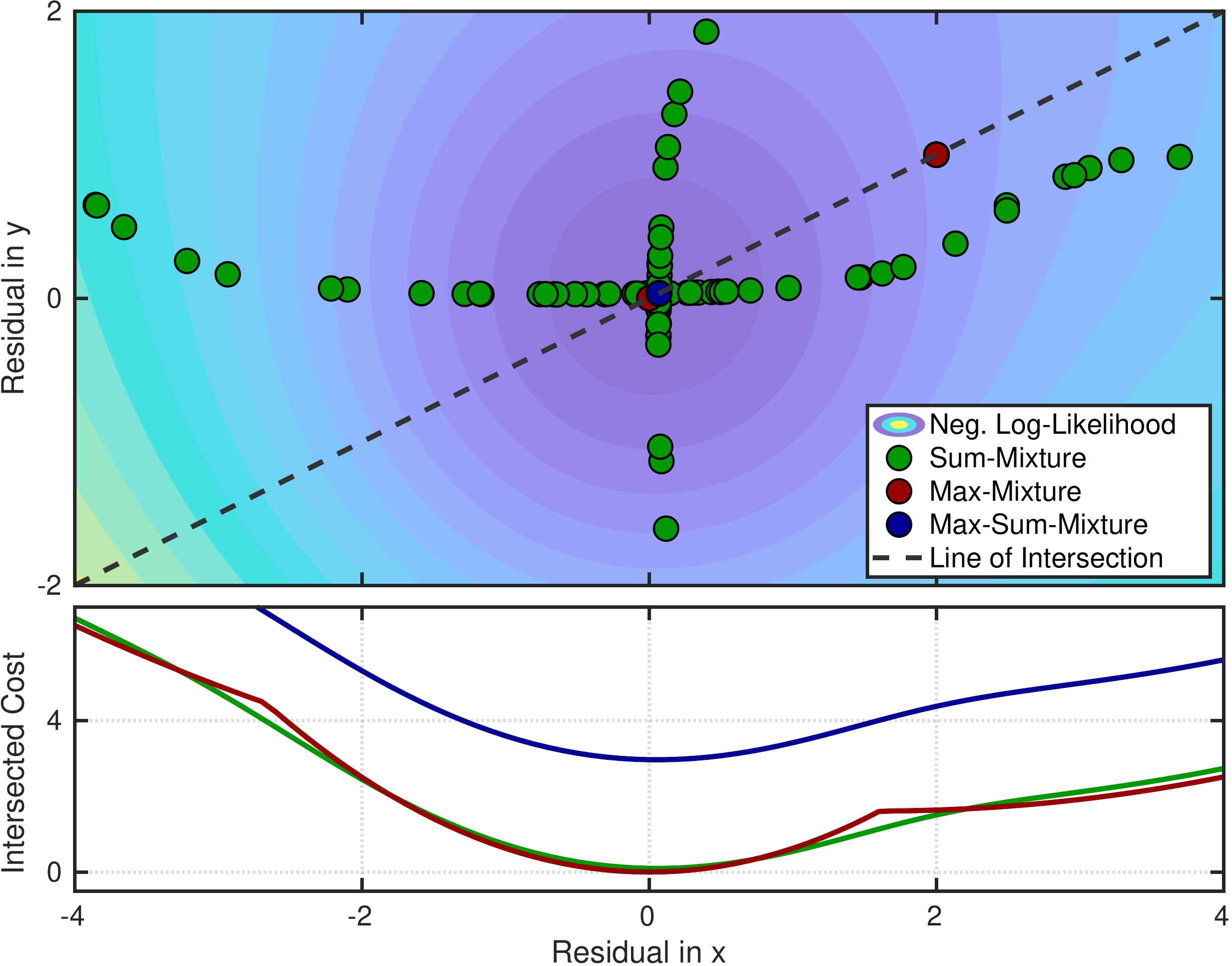}
    \caption
    {
        Optimization of a Gaussian mixture's negative log-likelihood.
        Starting from 100 different initial positions, a nonlinear least squares solver is applied to find the minimum of three cost function variants that represent the same mixture, as described in \autoref{sec:eval_plain}.
        The scattered points represent the results after each individual optimization.
        While the proposed Max-Sum-Mixture converges fully to the global minimum, Sum-Mixture fails to converge for almost all initial points.
        By intersecting the cost surface, the lower plot shows why some points of Max-Mixture get stuck: It has a second local minimum.
        Please notice, that both plots share the same x-axis.
    }
    \label{fig:error}
\end{figure}

In this paper, we introduce a novel representation for Gaussian mixtures in least squares problems, such as state estimation using factor graphs or point set registration.
Based on Sum-Mixture, we derive a hybrid formulation of Max-Mixture and Sum-Mixture using robust numerical methods in \autoref{sec:gmm}.
It offers the combined advantages of both, without their individual disadvantages.\clearpage
In addition, we provide:
\begin{itemize}
	\item A discussion how to reduce the impact of nonlinearity on the estimation process in \autoref{sec:damping}.
	\item A comparison against Max-Mixture and Sum-Mixture in extensive Monte Carlo experiments, as shown in \autoref{fig:error}, in \autoref{sec:eval_plain}.
	\item A novel method to apply the proposed model to point set registration in \autoref{sec:eval_psr}.
	\item A detailed analysis of the reasons why state-of-the-art approaches fail under certain conditions in \autoref{sec:diss}.
\end{itemize}
Furthermore, we provide open source implementations\footnote{Available under \url{www.mytuc.org/mix}} of our approach -- as well as SM and MM -- for two of the most common optimization frameworks in robotics: Ceres \cite{Agarwal} and GTSAM \cite{Dellaert}.

%% file: inc/prior.tex

\section{Prior Work} \label{sec:prior}

While the field of robust state estimation is remarkably broad, we focus on a brief overview of different robust probabilistic models for nonlinear least squares (NLS) and factor graph optimization.
The common goal of all algorithms is to reduce the huge impact of non-Gaussian outliers to least squares problems.

M-estimators are one of the standard solutions for robust estimation.
They describe a set of loss functions that weight large residuals down.
Frameworks like GTSAM or Ceres bring a variety of M-estimators with them, like the historical Huber M-estimator \cite{Huber1964} or the more recent Dynamic Covariance Scaling (DCS) \cite{Agarwal2013}.
Due to their symmetry, they are not suited to describe arbitrary multimodal distributions that arise from ambiguity or effects like reflected GNSS signals.
Also, their performance depends strongly on the choice of specific parameters, which often leads to manual tuning.
In more recent work like \cite{Agamennoni2015} and \cite{Chebrolu2021}, the parametrization issue was resolved by self-tuning algorithms based on expectation-maximization.
But still, they are limited by their symmetry constraint.

Switchable constraints \cite{Suenderhauf2012a} introduced weights for each residual that can be jointly optimized within the least squares problem.
In this way, the optimizer can down-weight outliers individually.
Albeit it is formulated different, the authors of \cite{Agarwal2013} showed that its solution is equivalent to an M-estimator.

Gaussian mixtures are quite flexible models but hard to integrate into nonlinear least squares.
So the authors of \cite{Pfingsthorn2012} proposed the idea of preselecting the dominant mode and using it as an approximation of the full GMM.
Since the optimization problem was constructed fully Gaussian afterwards, it is just a limited workaround and no reselection is possible.
Olson et al. went one step further in \cite{Olson2013} and proposed a better approximation -- a local maximum of multiple Gaussians.
In contrast to the explicit selection in \cite{Pfingsthorn2012}, the Max-Mixture approach selects the dominant mode locally and implicitly during the optimization.
MM is widely adopted in SLAM and factor graph optimization as an efficient and robust model.
However, it is still a limited approximation of the true density, especially GMMs with overlapping components are not represented well.  

With \cite{Rosen2013}, Rosen et al. proposed a very general approach to represent almost arbitrary distributions in NLS problems.
This idea had a fundamental importance for later work \cite{Pfeifer2016}, were we firstly implemented an exact GMM for NLS.
As we show in this publication, the Sum-Mixture model from \cite{Pfeifer2016} is fundamentally ill-posed beside some special cases.
Nevertheless, both papers \cite{Olson2013} and \cite{Rosen2013} inspired the novel approach that is presented in this work.

Besides least-squares optimization, the authors of \cite{Stoyanov12a} applied Gaussian mixtures to solve probabilistic point set registration.
They used the full Gaussian mixture, but approximated the product of probability densities by a sum of them, which violates fundamental assumptions of probability theory.

The authors of \cite{Hsiao2019} and \cite{Fourie2016} introduced novel algorithms to represent ambiguity, which includes equivalents of Gaussian mixture models.
However, they require different kinds of specialized solvers and come with a lot of overhead.

We, instead, present a novel Gaussian-mixture implementation for common nonlinear least squares solvers, that can be applied to a variety of robotic applications.
Our approach combines the well-behaved convergence of Max-Mixture with the exact representation of Sum-Mixture without adopting their particular weaknesses.

%% file: inc/gmm.tex

\begin{table*}[t]
	\centering
	\caption{Gaussian Mixture Loss Functions for Least Squares}
	\label{tab:cost}
	\begin{tabular}{lccc}
		\toprule
		Algorithm & Max-Mixture \cite{Olson2013} & Sum-Mixture \cite{Rosen2013} & Max-Sum-Mixture\\ \midrule
		Notation & $\lossMM$ & $\lossSM$ & $\lossMSM$\\ \midrule
		Cost Function
		&
		$
		\begin{bmatrix}[2]\displaystyle
			\sqrtinfo_\iMax \left( \residual - \mean_\iMax \right)\\\displaystyle
			\sqrt{- 2 \log \left(  \dfrac{\weight_\iMax \det \sqrtinfo_\iMax}{\normMM} \right) }
		\end{bmatrix}
		$
		& 
		$ \displaystyle
		\sqrt{-2\exponent_\iMax - 2 \log \left( \frac{1}{\normSM} \sumGMM \scaling_\iGMM \exp \left( \exponent_\iGMM - \exponent_\iMax \right)  \right)}
		$
		&
		$
		\begin{bmatrix}[2]\displaystyle
			\sqrtinfo_\iMax \left( \residual - \mean_\iMax \right)\\\displaystyle
			\sqrt{-2 \log \left(\frac{1}{\normMSM} \sumGMM \scaling_\iGMM \exp \left( \exponent_\iGMM - \exponent_\iMax \right)  \right)}
		\end{bmatrix}
		$
		\\ \midrule
		Normalization
		&
		 $\normMM \geq \maxGMM\weight_\iGMM \det \sqrtinfo_\iGMM$ 
		&
		 $\normSM \geq \sum_\iGMM \weight_\iGMM \det \sqrtinfo_\iGMM$
		&
		$\normMSM = \iGMMEnd \cdot \maxGMM \weight_\iGMM \det \sqrtinfo_\iGMM + \delta$ \\ \bottomrule
	\end{tabular}
\end{table*}

\section{Gaussian Mixture Models for Least Squares} \label{sec:gmm}

To introduce our novel Gaussian mixture representation for least squares, we take a look at the Gaussian case before we extend it to mixtures.

Central idea of probabilistic state estimation is to find the most likely set of states $\State$ based on a set of perceived measurements $\Measurement$.
The likelihood maximization is done by minimizing the negative log-likelihood, which leads to the common least squares formulation: 
\begin{align}
	\StateEst & = \argmax_{\State} \prob{\State \given \Measurement}                                                                      \\
	          & = \argmin_{\State} -\log \prob{\State \given \Measurement}                                                                \\
	          & = \argmin_{\State} -\sum_{n} \log \prob{\residual(\state_n, \measurement_n)}   				\label{eqn:log_like}          \\
	          & = \argmin_{\State} \frac{1}{2} \sum_{n} \left\| \residual(\state_n, \measurement_n) \right\|^{2}_{\cov} + C
\label{eqn:state_est}
\end{align}
Please note, that we use two notations for the squared Mahalanobis distance, which are
\begin{equation}
\big\| \residual \big\|^{2}_{\cov} = \big\| \sqrtinfo \residual  \big\|^{2},
\label{eqn:mahala}
\end{equation}
where $\cov$ is the measurement covariance matrix and $\sqrtinfo$ is the square-root information matrix.
Residual term $\residual$ describes a nonlinear difference between a subset of states $\state_n$ and measurements $\measurement_n$.
Constant $C$ is the Gaussian log-normalization and does not depend on the states $\State$, so it can be omitted.
The remaining pure quadratic form is the direct result of assuming Gaussian distributed noise for the residual $\residual$.
Since a quadratic problem is sensitive to any kind of outlier, the residual term is often wrapped into a loss function $\loss$ which brings us to robust least squares:
\begin{equation}
\StateEst  = \argmin_{\State} \frac{1}{2} \sum_{n} \left\| \loss(\residual(\state_n, \measurement_n)) \right\| ^2
\label{eqn:state_est_robust}
\end{equation}
This new cost function \autoref{eqn:state_est_robust} is composed of the residual function $\residual$ and the loss function $\loss$.
The loss can be derived directly from negative log-likelihood of the residual \autoref{eqn:log_like} to apply different probabilistic models:
\begin{equation}
\loss(\residual) = \sqrt{-2\log(\prob{\residual})}
\end{equation}
The Mahalanobis distance \autoref{eqn:mahala} is a special case of this loss function.
Please note, that we define $\loss$ inside the square of \autoref{eqn:state_est_robust} to allow asymmetric loss functions, which is slightly different to the literature.
The main contribution of this paper can be condensed down to the answer to a simple question:
\emph{How can we represent the log-likelihood of a GMM using the loss function $\loss(\residual)$?}

\subsection{Max-Sum-Mixture -- A Novel Approach}

For a better understanding of the underlying mathematical structure, we introduce a compressed notation for the probability density of a Gaussian mixture with $\iGMMEnd$ components:
\begin{equation}
\prob{\residual} \propto \sumGMM \scaling_\iGMM  \exp(\exponent_\iGMM(\residual))
\label{eqn:gmm_comp}
\end{equation}
The constant scaling $\scaling$ contains the weighting $\weight$ of a component $\iGMM$ and the square root information matrix $\sqrtinfo$:
\begin{equation}
\scaling_\iGMM = \weight_\iGMM  \det \sqrtinfo_\iGMM
\label{eqn:gmm_scale}
\end{equation}
The exponent $\exponent$ is defined as half of the squared Mahalanobis distance between the residual $\residual$ and the mean $\mean$:
\begin{equation}
\exponent_\iGMM(\residual) = - \frac{1}{2} \left\| \sqrtinfo_\iGMM \left( \residual - \mean_\iGMM \right) \right\| ^2 
\label{eqn:gmm_expo}
\end{equation}
The log-likelihood of \autoref{eqn:gmm_comp} is the scaled version of the log-sum-exp function
\begin{equation}
\lse(\vect{\scaling},\vect{\exponent}) = \log \left(\prob{\residual} \right)  =  \log \left( \sumGMM \scaling_\iGMM  \exp (\exponent_\iGMM) \right),
\label{eqn:log-sum-exp}
\end{equation}
with the set of scalings $\vect{\scaling} = \left\lbrace \scaling_1, \dots, \scaling_\iGMMEnd \right\rbrace$ and exponents $\vect{\exponent} = \left\lbrace \exponent_1, \dots, \exponent_\iGMMEnd \right\rbrace$.
While it is not possible to push the logarithm inside the sum, it is indeed possible to pull one particular exponent $\exponent_\iMax$ outside:
\begin{align}
-\lse(\vect{\scaling},\vect{\exponent})
&= -\log \left( \sumGMM \scaling_\iGMM  \exp (\exponent_\iGMM) \right)\nonumber\\
&= -\log \left( \sumGMM \frac{\scaling_\iGMM  \exp (\exponent_\iGMM)}{ \exp (\exponent_\iMax)}  \cdot \exp (\exponent_\iMax) \right)\nonumber\\
&= -\underbrace{\exponent_\iMax \vphantom{\sumGMM}}_{\text{linear}}
- \underbrace{\log \left( \sumGMM \scaling_\iGMM \exp (\exponent_\iGMM -\exponent_\iMax) \right)}_{\text{nonlinear}} \label{eqn:msm_detail}
\end{align}
This is a common approach to formulate a numerical robust log-sum-exp function, but we apply it algebraically to split the sum in a linear and a remaining nonlinear part.
Since $\exponent_\iMax$ can be chosen freely, we choose the locally dominant component to get a big linear term.
So the index $\iMax$ of $\exponent_\iMax$ is defined by
\begin{equation}
\iMax = \argmaxGMM \left(  \scaling_\iGMM \exp (\exponent_\iGMM) \right).
\label{eqn:msm_exp}
\end{equation}
Interestingly, this part is identical with the Max-Mixture formulation from \cite{Olson2013}.
In difference to MM, we keep the remaining nonlinear term in the least squares problem.
Therefore, we redefine the set of exponents as
\begin{equation}
	\vect{\tilde{\exponent}} = \left\lbrace \exponent_1 - \exponent_\iMax, \dots, \exponent_\iGMMEnd - \exponent_\iMax \right\rbrace,
\end{equation}
and apply the generalization approach of \cite{Rosen2013} by introducing a normalization constant $\normMSM$:
\begin{equation}
-\lse(\vect{\scaling}, \vect{\tilde{\exponent}})
\propto \half \left\|  
	\sqrt{-2 \lse \left( \frac{\vect{\scaling}}{\normMSM},\vect{\tilde{\exponent}} \right) }
		\right\| ^2 
\label{eqn:msm_nonlin}
\end{equation}
The normalization guarantees that the expression inside the square root is always positive.
As shown in detail in Appendix \ref{app:proof}, it can be set to
\begin{equation}
\normMSM = \iGMMEnd \cdot \maxGMM \scaling_\iGMM + \delta \text{ with } \delta \geq 0.
\label{eqn:msm_norm}
\end{equation}
Damping parameter $\delta$ can be used to reduce the nonlinearity of \autoref{eqn:msm_nonlin} -- its optimal choice is discussed the next section.
Now, everything can be put together to formulate our main contribution -- the square root of the negative log-likelihood of a GMM:
\begin{equation}
\lossMSM(\residual) = 
\begin{bmatrix}[1.5]\displaystyle
\sqrtinfo_\iMax \left( \residual - \mean_\iMax \right)\\\displaystyle
\sqrt{-2 \log \left(\frac{1}{\normMSM} \sumGMM \scaling_\iGMM \exp \left( \exponent_\iGMM - \exponent_\iMax \right)  \right)}
\end{bmatrix}
\label{eqn:msm_final}
\end{equation}
Please note that this is a vector expression with a dimensionality of $\dim(\residual) + 1$.
The related Jacobian is provided in Appendix \ref{app:jacobian}.

\begin{figure*}[tbph]
	\centering
	\begin{minipage}[b]{0.485\textwidth}
		\centering
		\includegraphics[width=0.95\linewidth]{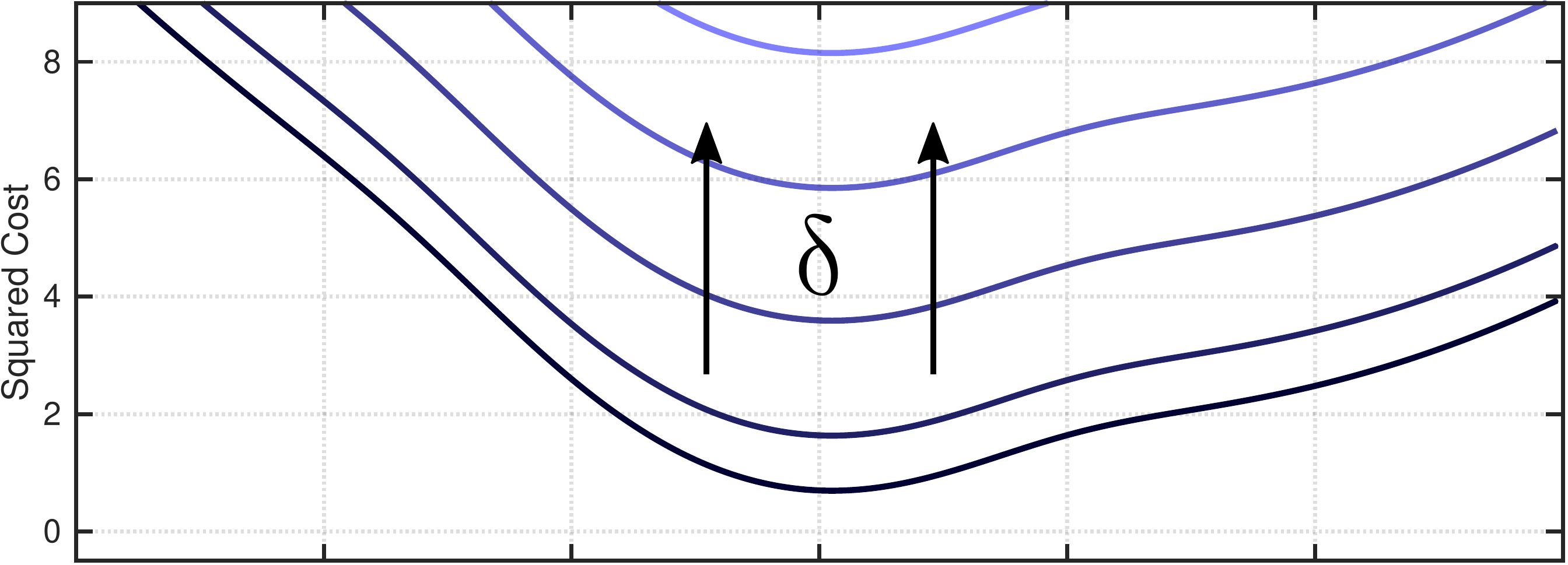}\\%
		\vskip 4pt
		\includegraphics[width=0.95\linewidth]{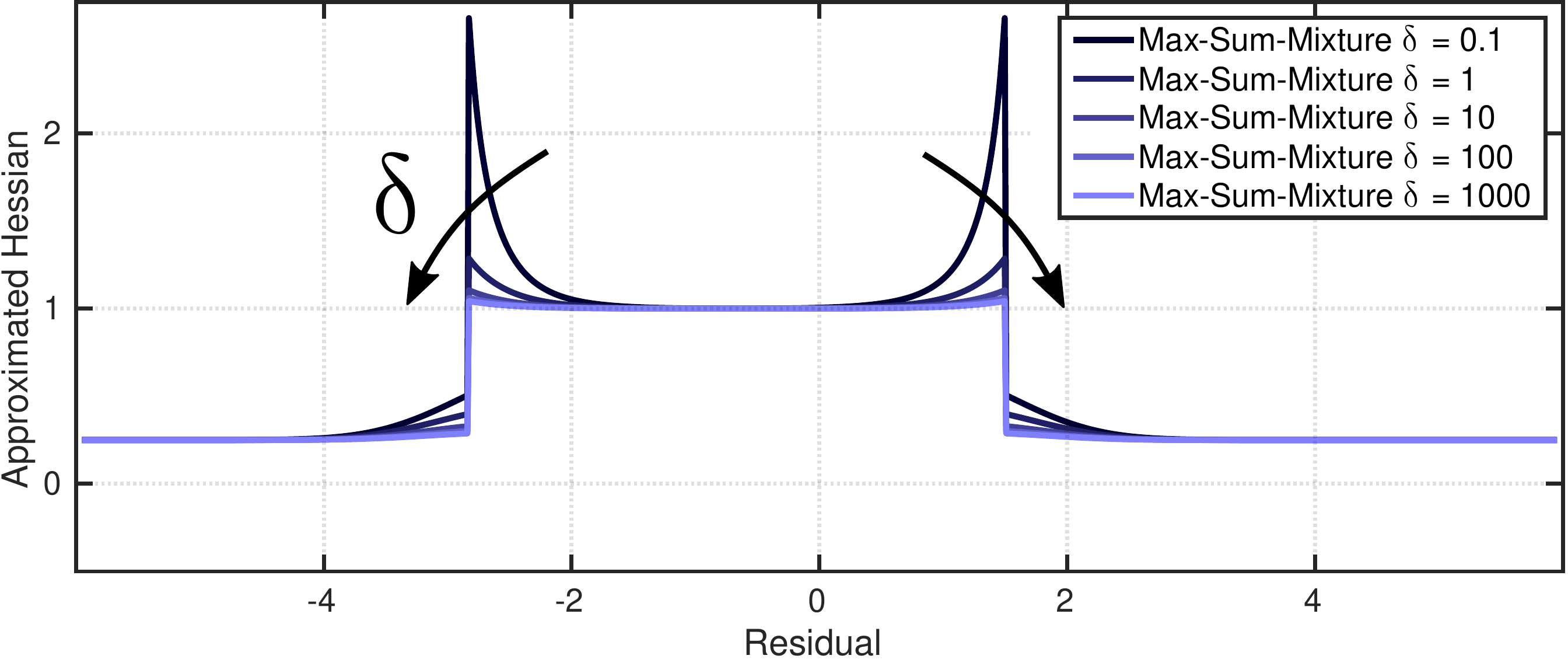}
		\caption
		{
			The influence of the damping factor $\delta$ on the squared cost and the approximated Hessian of our approach.
			Each line represents a Max-Sum-Mixture cost function with a different value of $\delta$ -- as darker the line, as lower the damping.
			A higher damping decreases the effect of nonlinearities on the Hessian, but shifts its cost surface upwards.
		}
		\label{fig:com_damping}
	\end{minipage}
	\hfill
	\begin{minipage}[b]{0.485\textwidth}
		\centering
		\includegraphics[width=0.95\linewidth]{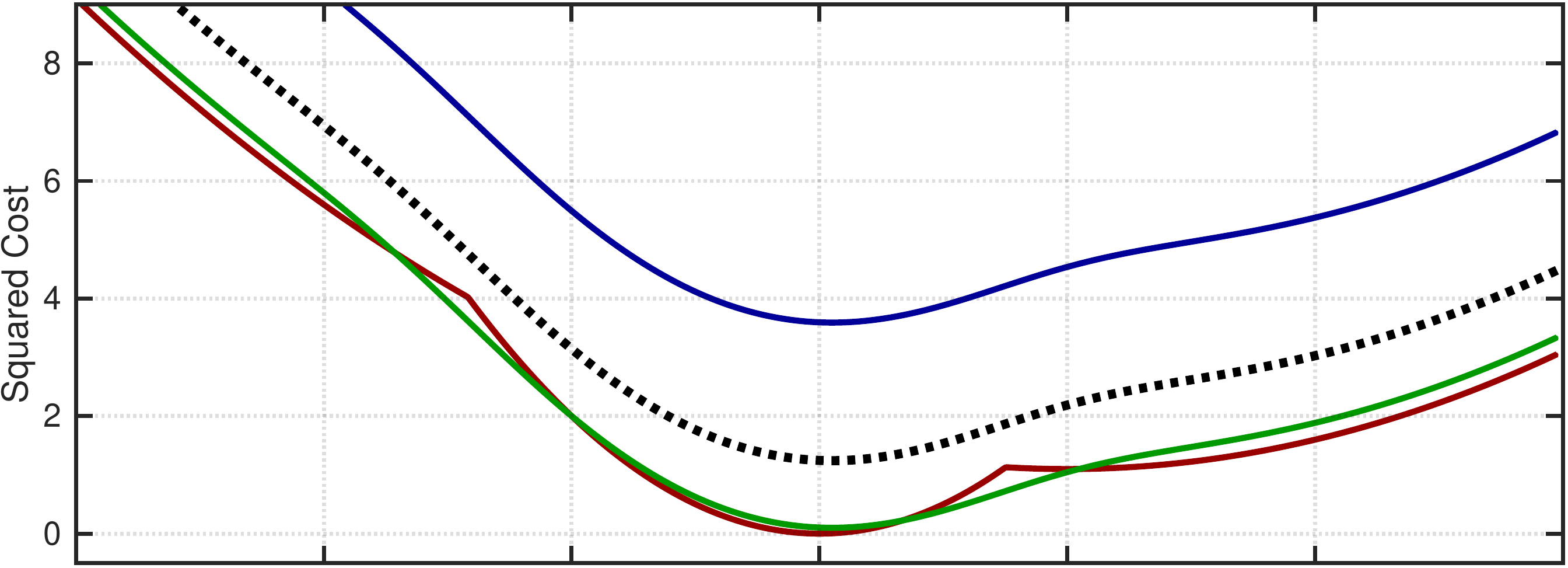}\\%
		\vskip 4pt
		\includegraphics[width=0.95\linewidth]{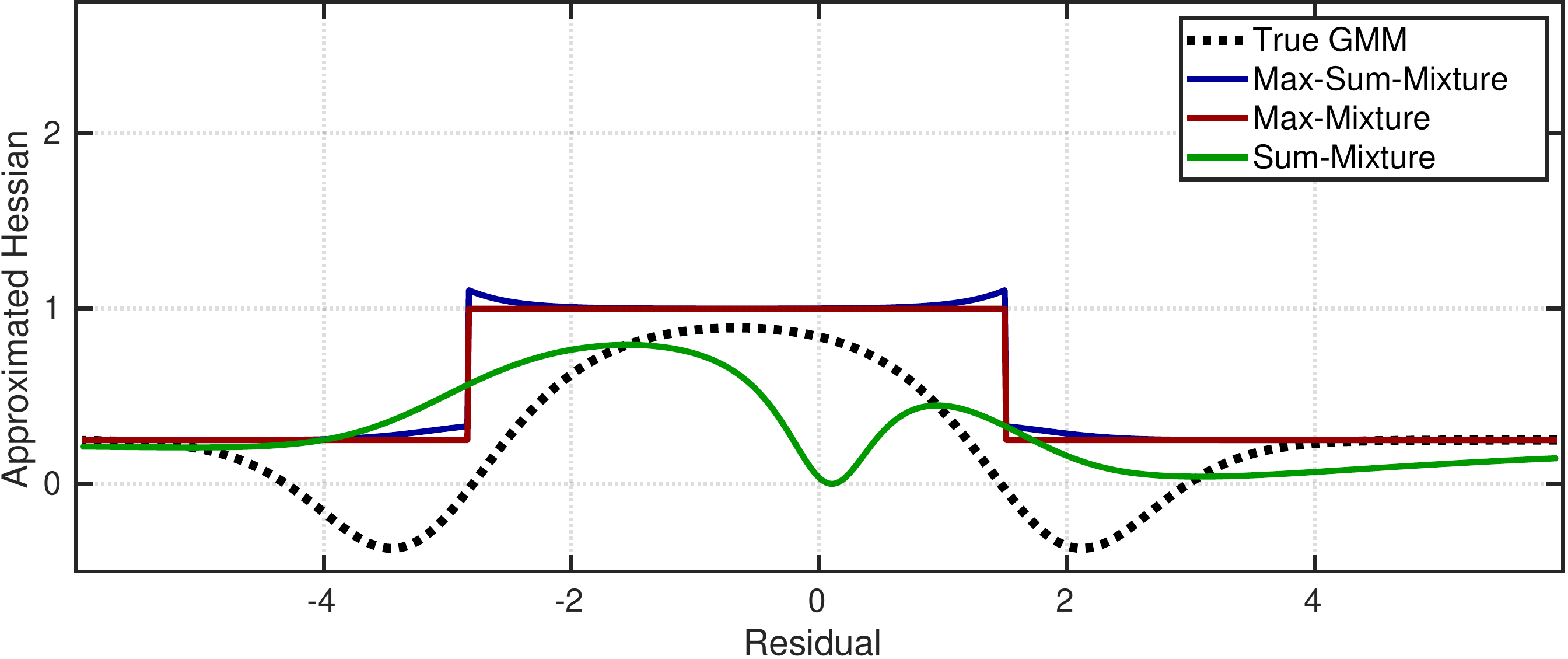}
		\caption
		{
			Comparison of cost functions and approximated Hessians of different mixture representations against the true Gaussian mixture.
			The dotted line represents the negative log-likelihood of the underlying GMM and its true Hessian.	
			Solid lines represent the different approaches to describe this GMM for least squares.
			The Hessians in the lower plot are approximated by squared Jacobians.
			At the cost minimum, Max-Sum-Mixture and Max-Mixture are close to the true Hessian, while the Hessian of Sum-Mixture is singular.
		}
		\label{fig:com_algo}
	\end{minipage}
\end{figure*}

\subsection{Damping the Nonlinearity} \label{sec:damping}

The previously introduced damping factor $\delta$ seems superfluous in the first place, but it can be used to control the impact of nonlinearities.
Therefore, its worth thinking about the best choice of $\delta$.

The effect of the damping factor on the optimization itself is rather small, since it only shifts the cost surface.
However, as elaborated in \cite[p.~22]{Madsen2004}, shifting the cost can decrease the rate of convergence.
Additionally, a high $\delta$ deceases the relative cost change, which can lead to an early termination of the optimization.
Therefore, $\delta$ should not be too high.

The damping factor affects also the estimated covariance $\covEst_\state$ of a state $\state$, that can be recovered from the optimization problem.
Covariance recovery can be done by inverting the Hessian $\mat{H}$ that is approximated by the squared Jacobian $\mat{J}$:
\begin{equation}
\covEst_\State = \mat{H}^{-1} \approx (\mat{J}\T \mat{J})^{-1}
\label{eqn:cov_recovery}
\end{equation}
We refer to this approximated Hessian as pseudo-Hessian.
In \autoref{fig:com_damping}, we illustrate the result of this approximation for different choices of the damping parameter.
For a high $\delta$, the pseudo-Hessian of MSM becomes similar to the pseudo-Hessian of MM, which is slightly to big -- and therefore slightly overconfident as shown in \autoref{fig:com_algo}.
As smaller the damping gets, as more overconfident the pseudo-Hessian becomes.
So, $\delta$ should not be too low.

Based on practical experience, almost any value will work for the damping factor.
Nevertheless, we could achieve the best results with values of between 1 and 1000.
So, we recommend using $\delta = 10$ and apply this for our evaluation.
\autoref{fig:com_algo} shows the resulting cost function compared to Max-Mixture \cite{Olson2013} and Sum-Mixture \cite{Rosen2013}.

\subsection{Relation between Sum-Mixture and Max-Sum-Mixture}

Since the proposed Max-Sum-Mixture has a substantial algebraic overlap with Sum-Mixture, we take a closer look at their mathematical relation.
Both loss functions are provided together with MM in \autoref{tab:cost} and their Jacobians are given in Appendix \ref{app:jacobian}.
For a fair evaluation, we apply the numerical robust log-sum-exp formulation for SM, even though this was not part of the original publication.

By comparing the scalar loss $\lossSM$ of SM with the intermediate step \autoref{eqn:msm_detail} of MSM, it becomes clear that both describe the same log-likelihood and their squared costs are directly proportional:
\begin{equation}
\lossSM^2 \propto \lossMSM\T \cdot \lossMSM
\end{equation}
They differ only in two points: their normalization constants are slightly different and MSM uses a vectorized loss function where the dominant component is pulled out.

Since the normalization constants barely influence the optimization, the major advantage of MSM stems from the vectorization.
Separating the dominant linear part from the nonlinear in \autoref{eqn:msm_final}, exposes a linear representation of the dominant mode to the optimizer.
For SM on the other hand, the optimizer has to calculate its steps based on the Jacobian of a completely nonlinear cost term.
Thus, the actual steps of MSM behave better than those of SM, even though their squared costs are almost equal.
In consequence, the proposed MSM is a much more efficient implementation of the concepts of \cite{Rosen2013} than the original scalar formulation.
We will verify this advantage in the next section.

%% file: inc/eval.tex

\section{Monte Carlo Evaluation} \label{sec:eval}

In the following, we explain the motivation behind our simulative evaluation.
Our experiments are designed to analyze the properties of the proposed Max-Sum-Mixture formulation compared to state-of-the-art approaches from \cite{Olson2013} and \cite{Rosen2013}.
A good probabilistic loss function should be an accurate and efficient representation of the corresponding distribution.
Therefore, we stress all approaches for their convergence properties, their accuracy, their credibility and their computational efficiency.

We provide two types of synthetic Monte Carlo experiments to generalize our findings beyond specific applications.
A plain optimization of a random set of Gaussian mixtures and a set of randomized point set registration problems.
All experiments were executed on a standard AMD Ryzen 7 desktop PC.

\subsection{Plain Optimization} \label{sec:eval_plain}
We design the first synthetic experiment to test our approach for a well-behaved convergence. 
This includes the ability to represent the true mode of a Gaussian mixture and a cost surface that allows the optimizer to converge fast.  

Hence, we create the simplest possible optimization problem that involves a GMM -- a single linear residual wrapped into a nonlinear loss function:
\begin{equation}
\stateEst = \argmin_{\state} \half \left\| \loss(\state) \right\| ^2
\end{equation}
The loss function represents the negative log-likelihood as described in \autoref{eqn:msm_final}.
To allow a rough comparison with M-estimators, we also add DCS \cite{Agarwal2013} to the symmetric scenarios.
Starting from \num{100} different initial positions, least-squares optimization is applied to find the minimum of each cost function.
By repeating the experiment for a set of \num{1000} different Gaussian mixtures, we run \num{100000} optimizations in total.

\subsubsection*{Dataset Generation}
To generate a variety of different models, we sample the mixture parameters randomly from a set of uniform distributions.
All mixtures are composed of two components, the first one with a zero mean and a standard deviation drawn between \SI{0.1}{\meter} and \SI{1}{\meter}.
The second component has a random mean from the range of \SI{\pm2}{\meter} and the standard deviation of the first, multiplied by a random factor between \num{2} and \num{10}.
A weight between \num{0.2} and \num{0.8} is used for the first component -- the second weight is simply the difference to \num{1}.
For two-dimensional models, we draw the values for both dimensions independently.
To guarantee that converging to the global optimum is always possible, we detect models with multiple minima and reject them, using a combination of grid-based sampling and local maxima detection.
The initial positions are linearly spaced in a range of \SI{\pm4}{\meter} to cover the relevant cost surface.
We implement the evaluation as part of the libRSF \cite{Pfeifer}, using the Ceres \cite{Agarwal} version of the Levenberg-Marquardt algorithm.

\subsubsection*{Metrics}
We evaluate the root-mean-square error (RMSE) as distance between the true mode of the GMM and the optimized variable after convergence.
The true mode is found by using a grid search, followed by a local optimization.
To check if an optimization converged to the right minimum, we apply an error threshold of \SI{0.01}{\meter} and define the ratio of runs below this limit as convergence success rate.
All results are listed together with the averaged run time and number of iterations in \autoref{tab:result_single_cost}.

\subsection{Point Set Registration} \label{sec:eval_psr}

Our motivation behind the second simulated evaluation stems from ego-motion estimation using sensors that measure sets of spatial points, such as radars or lidars.
This task can be interpreted as a robust point set registration problem, as shown for example by \cite{Barjenbruch15} with radar sensors.
We apply the concept from \cite{Jian11} to represent two set of points with Gaussian mixture models and register them in a distribution-to-distribution manner.
By using a GMM representation, we can simply integrate a varying uncertainty of each target point (or sometimes referred to as landmark) within the two point sets -- crucial for having a credible estimator.
Furthermore, we can omit the usually required step of finding explicit correspondences compared to other point set registration algorithms.

However, with focus on credibility, we differ from the implementation in \cite{Barjenbruch15}.
They used a sum approximation to align both GMM distributions which leads to robustness against outliers, but results in wrong covariances.
Their approach increases furthermore the possible number of local minima -- we refer to future work for a more detailed investigation.

\subsubsection*{Theoretical Background}
In general, point set registration is the task of finding the transformation which aligns a current or moving point set $\cM$, so that it overlaps with a reference or fixed set $\cF$. We use the index $i$ for a point measurement $\vm_i$ within $\cM$ and $j$ for a point in $\cF$ respectively. For finding the best transformation $\vx =  \left[ \vx^{\text{tran}},  \vx^{\text{rot}} \right] \T$, we can maximize the joint distribution
\begin{align}
    P(\cM | \cF, \vx) = \prod_{i=1}^{\left|\cM\right|} P\left( \vm_i \given \cF, \vx\right).
\end{align}
Following the idea of \cite{Jian11} and \cite{Stoyanov12a}, we model the reference set as GMM, which results in the following distribution for one point $\vm_i$ of $\cM$:
\begin{align}
    P(\vm_i | \cF) = \sum_{j=1}^{\left| \cF \right|} w \cdot \Normal{\vm_i \given \vmu_j, \vSigma_j} \text{ with } w = \frac{1}{\left| \cF \right|}
\end{align}
Taking into account the transformation and rewriting the normalization terms leads to the likelihood function
\begin{align}
    P(\vm_i | \cF, \vx) \propto \sum_{j=1}^{\left| \cF \right|} \scaling_j \exp \left( -\frac{1}{2} \smd{\vr_i - \vmu_j}{\vSigma_j+\tilde\vSigma_i} \right).
    \label{eqn:gmm_psr}
\end{align}
The residual $\vr_i$ represents the transformation of point $\vm_i$ into the reference frame with the current estimate of $\vx$:
\begin{align}
    \vr_i (\vx) = \vR(\vx^{\text{rot}}) \vm_i + \vx^{\text{tran}}
\end{align}
Also, the covariance matrix $\tilde\vSigma_i = \vR \vSigma_i \vR\T$ of $\vm_i$ is transformed into the current frame, using the current estimate's rotation matrix $\vR(\vx^{\text{rot}})$.
The resulting least squares problem is
\begin{equation}
	\stateEst = \argmin_{\state} \sum_{i=1}^{\left|\cM\right|} \half \left\| \loss(\vr_i(\vx)) \right\| ^2,
\end{equation}
where the loss $\loss$ describes the negative log-likelihood of the GMM \autoref{eqn:gmm_psr}.
Note, that we can simply add an uncertain Gaussian component to the fixed frame, representing outliers, to have a robust version of the point set registration problem.
We use this robust version in our last experiment with the Manhattan dataset.
Again, Levenberg-Marquardt is applied to solve the optimization, in this experiments based on the GTSAM framework \cite{Dellaert}.

\begin{figure}[tbp]
	\centering
	\includegraphics[width=0.95\linewidth]{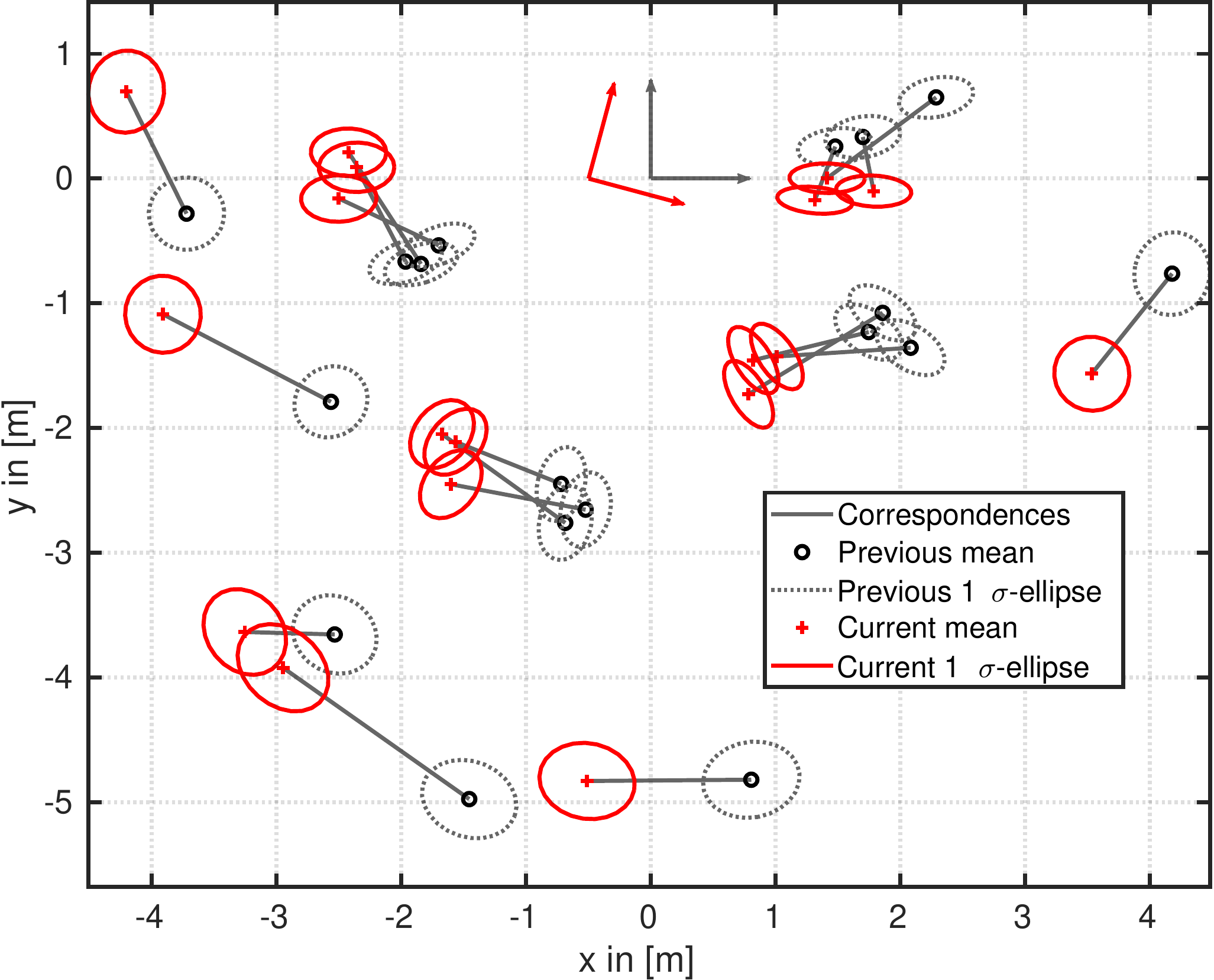}
	\caption
	{
		Example of one random point set registration problem with otherwise unknown correspondence information for better visual association.
		The transformation of the ground truth points, before adding simulated measurement noise to both sets, was \SI{0.5}{\metre} in x-direction and \SI{15}{\degree} in rotation.
		Specific properties of the described experiment like a measurement noise based on polar coordinates and clustered landmarks are apparent.  
	}
	\label{fig:psr2dProblem}
\end{figure}

\subsubsection*{Dataset Generation} 
\begin{table}[tbp]
	\centering
	\caption{Parameters for the Point Set Registration Experiments.}
	\label{tab:parameters}
	\sisetup{round-mode = off,separate-uncertainty = true}
	\begin{tabular}{@{}m{2.9cm}cc@{}}
		\toprule
		\textbf{Experiment}         & \textbf{2D} & \textbf{3D} \\ \midrule
		Num. of Configurations      & 100 & 100 \\%
		\arrayrulecolor{lightgray}\midrule 
		Landmark\newline Generation & 
		$\begin{aligned} &\uniform{x}{\SI{+-5}{\metre}}      \\ &\uniform{y}{\SI{+-5}{\metre}}\end{aligned}$ & 
		$\begin{aligned} &\uniform{r}{\SI{10+-1}{\metre}} \\ &\uniform{\theta}{\num{+-\pi}}\\ &\uniform{\phi}{\pm\num[fraction-function = \tfrac]{1 / 2} \pi}\end{aligned}$ \\%
		\midrule
		Clustering (size)           & \SI{40}{\percent} (\num{3}) & \SI{40}{\percent} (\num{3}) \\ 
		Cluster spread              & $\normal{x,y}{0,0.1^2}$ & $\normal{x,y,z}{0,0.1^2}$ \\%
		\arrayrulecolor{black}\midrule
		Runs per Configuration      & 1000 & 1000 \\%
		\arrayrulecolor{lightgray}\midrule
		Transformation\newline sampling &
		$\begin{aligned} &\uniform{x,y}{\SI{+-0.5}{\metre}}   \\ &\uniform{\alpha}{\ang{+-15}}\end{aligned}$ & 
		$\begin{aligned} &\uniform{x,y,z}{\SI{+-0.5}{\metre}} \\ &\uniform{\alpha_{x,y,z}}{\ang{+-5}}\end{aligned}$\\%
		\arrayrulecolor{lightgray}\midrule
		Measurement noise           &
		$\begin{aligned} &\normal{r}{0,\left(\SI{0.2}{\metre}\right)^2}\\ &\normal{\theta}{0,\left(\ang{3}\right)^2}\end{aligned}$ &
		$\begin{aligned} &\normal{r}{0,\left(\SI{0.2}{\metre}\right)^2}\\ &\normal{\theta,\phi}{0,\left(\ang{3}\right)^2}\end{aligned}$ \\%
		\arrayrulecolor{black}\bottomrule
	\end{tabular}
\end{table}

To evaluate the different mixture models applied to the point set registration problem, we simulate \num{1000} different measurement pairs on \num{100} different landmark configurations. 
Each consisting of \num{10} uniformly distributed landmarks in a square of \SI[product-units = brackets]{10 x 10}{\metre}.
Following the typical characteristics of an automotive radar sensor, \SI{40}{\percent} of the landmarks are duplicated two times based on a standard deviation of \SI{0.1}{\metre} to generate clusters. 
Now, we assign a measurement noise (in polar space) with standard deviation of \ang{3} and \SI{0.2}{\metre} to  each of the \num{18} landmarks.
Based on this, we sample a new measurement pair for each Monte Carlo run. 
The transformation between the measurements follows a uniform distribution in the range of \SI{\pm 0.5}{\metre} in x- and y-direction and \ang{+-15} for the angle. 
The random set of \num{1000} different transformations is the same for each of the \num{100} landmark configurations.

In addition to the 2D dataset, a 3D variant with 6 degrees of freedom was generated in the same style, but with \num{20} landmarks to compensate for the higher degree of freedom. 
Parameters for both experiments are summarized in \autoref{tab:parameters}, where the variables $r$, $\theta$, and ($\phi$) are used for radius, azimuth, and (elevation) to define parameters in polar and (spherical) coordinates.
Please note, that $r$ means the radius in this context, while $\residual$ is the residual function.
Furthermore, we use Cartesian coordinates with the nomenclature $x,y,z$ and an Euler angle representation with $\alpha_{x,y,z}$ for rotations around the corresponding axes. 

Since real world applications have to deal with outliers and sets of not fully overlapping points, we include an additional evaluation based on the Manhattan dataset M3500b from \cite{Carlone14}.
A number of \num{300} landmarks are added uniformly into the world with the same clustering and noise properties as before.
Instead of random transformations, the dataset's trajectory is used -- excluding loop closure transitions.
Landmark measurements are simulated by a sensor with \SI{10}{\metre} range and \SI{320}{\degree} FoV, resulting in an average outlier rate of \SI{13}{\percent} and a maximum of \SI{32}{\percent}.%

\subsubsection*{Metrics}
As it is common practice, we apply the RMSE separately for the translational and rotational errors to measure accuracy.
Aside from accuracy, we also want to measure, how good the estimator calculates its own uncertainty -- also referred to as the estimator's credibility.
Therefore, the average normalized estimation error squared (ANEES) is used as originally described in \cite[Chp.~5.4]{Shalom01}, but normalized with the problem's dimension as stated in \cite{li02}.
If the estimator is perfectly credible, the ANEES value equals \num{1}.
Since the optimization problem is likely to have multiple local minima, we expect that none of the evaluated algorithms will be fully credible.
Nevertheless, we use the ANEES as a comparative metric to evaluate which algorithm is the most credible under this rough conditions.

\begin{table}[tbp]
	\centering
	\caption{Results of the Single Cost Function Evaluation.}
	\label{tab:result_single_cost}
	\sisetup{output-exponent-marker=\ensuremath{\mathrm{e}},
			 exponent-product={},
			 round-mode = places,
			 quotient-mode = fraction}
	\setlength{\tabcolsep}{5pt}
	\begin{tabular}{@{}l
					   l
					   S[table-format = 1.2e2, round-precision = 2, table-number-alignment = right]%
					   S[table-format = 3.1, round-precision = 1, table-number-alignment = center]%
					   S[table-format = 2.1, round-precision = 1, table-number-alignment = center]%
					   S[table-format = 3.0, round-precision = 0, table-number-alignment = center]%
					@{}}
		\toprule%
		 &	Alg.%
		 & {\begin{tabular}[c]{@{}c@{}} RMSE \\ \hspace{0pt}[\si{\meter}] \end{tabular}}%
		 & {\begin{tabular}[c]{@{}c@{}} Converg. Succ. \\ Rate [\SI{}{\percent}]\end{tabular}}%
		 & {\begin{tabular}[c]{@{}c@{}} Average \\ Iterations\end{tabular}}%
		 & {\begin{tabular}[c]{@{}c@{}} Average \\ Runtime [\SI{}{\micro\second}]\end{tabular}}	\\ \midrule
		\multirow{4}{*}{\begin{tabular}[c]{@{}c@{}}1D\\ sym.\end{tabular}}  & DCS  		& 7.4759e-14 	& 100.0  & 4.1715  & 25.642  \\
																			& MM  		& 9.636e-11 	& 100.0  & 3.9572  & 25.35   \\
																			& SM     	& 4.7539e-09 	& 100.0  & 28.986  & 154.93 \\
																			& MSM		& 8.8309e-05 	& 100.0  & 6.979   & 44.21  \\ \midrule
		\multirow{3}{*}{\begin{tabular}[c]{@{}c@{}}1D\\ asym.\end{tabular}} & MM  		& 0.71525 		& 60.1   & 3.4597  & 23.384  \\
																			& SM   		& 1.6521e-05 	& 100.0  & 29.283  & 141.98 \\
																			& MSM 		& 9.466e-05 	& 100.0  & 9.5919  & 57.807  \\ \midrule
		\multirow{4}{*}{\begin{tabular}[c]{@{}c@{}}2D\\ sym.\end{tabular}}  & DCS  		& 1.3371e-13 	& 100.0  & 4       & 25.214  \\
																			& MM   		& 3.3797e-07 	& 100.0  & 3.8912  & 25.55  \\
																			& SM      	& 2.1877 		& 6.82   & 75.724  & 414.52 \\
																			& MSM 		& 5.1120e-04 	& 100.0  & 3.5546  & 26.563  \\ \midrule
		\multirow{3}{*}{\begin{tabular}[c]{@{}c@{}}2D\\ asym.\end{tabular}} & MM   		& 1.0891 		& 56.8   & 3.1098  & 22.66  \\
																			& SM      	& 2.3394 		& 3.75   & 75.222  & 407.31 \\
																			& MSM	 	& 5.9761e-04 	& 100.0  & 8.0358  & 50.868  \\ \bottomrule
	\end{tabular}
\end{table}

\newcommand{\splitcell}[1]{\begin{tabular}[c]{@{}c@{}}#1\end{tabular}}
\begin{table}[tbp]
	\centering
	\caption{Results of the Point Set Registration Experiments.}
	\label{tab:result}
	\setlength{\tabcolsep}{4pt}
	\sisetup{round-mode = places, round-precision = 2}
	\begin{tabular}{@{}
					  c
					  r
					  S[table-format=1.3, round-precision = 3]
					  S[table-format=1.2]
					  S[table-format=1.2]
					  S[table-format = 2.1,round-precision = 1]
					  S[table-format = 2.1,round-precision = 1,table-number-alignment = center]
					  @{}}
		\toprule
		&
		\multicolumn{1}{c}{Alg.} & 
		{\splitcell{RMSE\\ \hspace{0pt}[\si{\metre}]}} & 
		{\splitcell{RMSE\\ \hspace{0pt}[\si{\degree}]}} & 
		{ANEES} & 
		{\splitcell{Average\\ Iterations}} &
		{\splitcell{Average\\ Runtime [\si{\milli\second}]}} \\ \midrule
	    \multirow{3}{*}{\begin{tabular}[c]{@{}c@{}}2D\end{tabular}}  & MM    &  0.117 & 1.481 & 1.985 & 3.7 &  2.65 \\
		& SM    &  0.100 & 1.294 & 0.127 & 22.0 & 7.28 \\
		& MSM &  0.098 & 1.226 & 1.318 & 4.5 &  3.31 \\ \midrule
	    \multirow{3}{*}{\begin{tabular}[c]{@{}c@{}}3D\end{tabular}}  & MM    &  0.140 & 1.984 & 2.982 & 3.8 &  5.99 \\
		& SM    &  0.228 & 2.212 & 0.691 & 26.8 & 31.07 \\
		& MSM &  0.122 & 1.714 & 2.180 & 4.4 &  8.48 \\ \midrule
		\multirow{3}{*}{\begin{tabular}[c]{@{}c@{}}M3500b\\(2D)\end{tabular}}  & MM    &  0.156 & 2.104 & 2.092 & 3.4 & 3.714 \\
		& SM    &  0.178 & 2.217 & 0.400 & 19.4 & 7.429 \\
		& MSM &  0.132 & 1.290 & 1.374 & 5.2 & 4.000 \\
		\bottomrule
	\end{tabular}
\end{table}

%% file: inc/discussion.tex

\section{Discussion} \label{sec:diss}

Based on the Monte Carlo experiments, we discuss the performance of our approach in the following section -- for the numbers see \autoref{tab:result_single_cost} and \autoref{tab:result}.

In terms of accuracy, we have to distinguish between plain optimization and point set registration.
In our plain optimization evaluation, we can see that the accuracy depends strongly on a well-behaved convergence.
Due to its linearity, \emph{Max-Mixture} converges perfectly for symmetric mixture models.
However, it fails in almost \SI{40}{\percent} of the asymmetric cases, where it converges to a false local minimum.
Although we only include GMMs with a distinct global optimum, the MM approximation is prone to have multiple ones as shown in \autoref{fig:com_algo}.
This demonstrates that albeit it is a close approximation, it behaves differently from the exact model.
\emph{Sum-Mixture} on the other hand is an exact model and performs well for one-dimensional problems.
For two-dimensional problems, however, it fails as shown in \autoref{fig:error} on the first page.
A convergence success rate below \SI{10}{\percent} indicates that the underlying optimization problem is seriously ill-posed.
This is caused by the problematic Jacobian that drops to zero, as shown in \autoref{fig:com_algo}.
We cannot discuss this deeply here, but these singularities slow down the convergence in 1D and prevent it in 2D.
The proposed \emph{Max-Sum-Mixture} has none of these issues.
It converges fast to the correct minimum in all evaluated cases and achieves a consistently low RMSE.
The error is slightly higher than for well-converging cases of SM/MM, because of the early termination due to the remaining constant part of the cost function (see \autoref{sec:damping}).
However, with a magnitude of $10^{-5}$, this effect is usually negligible for practical applications.
The M-estimator DCS shows the highest accuracy through all algorithms, which is expectable, since it is perfectly quadratic in the final phase of convergence. Still, it is not applicable for asymmetric problems.

In difference to the artificial example of plain optimization, all approaches are able to converge in our point set registration experiment.
This is possible because the reasons of failure -- singularities of SM or minima of MM -- occur just locally in one of many cost terms.
To give an example: A cost function with a Jacobian of zero, does not contribute to the optimization if the other derivatives are non-zero -- the algorithm can still converge.
Nevertheless, the individual weaknesses of each approach degrade the accuracy and the proposed MSM provides the smallest RMSE in translation as well as in rotation.
SM achieves a similar accuracy in the 2D Monte Carlo scenario, but for 3D it has twice the error of MSM.
MM is more consistent than SM through the different scenarios, but its RMSE is about \SIrange[range-phrase=--, range-units=single]{15}{20}{\percent} higher than our proposed approach.
The comparatively low rotational error of MSM in the Manhattan world scenario supports these findings.
Since variant M3500b is characterized by a large initial orientation error, MSM can profit from its well-behaved convergence. 

Analyzing the credibility of the point set registration problem, shows again the consequences of the problematic Jacobian of SM -- a strongly underconfident result.
It is worth mentioning, that the ANEES is an inverse proportional value, so an ANEES of \num{0.13} means the estimated covariance is about \num{8} times higher than the actual squared error.
Hence, it is clear that in the 2D case, the covariance estimated by SM is way too high for practical applications.
MSM is just slightly overconfident with an ANEES of \num{1.3}, which is a good result if we take the multiple minima into account that occur in point set registration.
The estimated uncertainty of MM is similar to MSM (as shown by \autoref{fig:com_algo}) and since its error is higher, it is more optimistic (overconfident).
For 3D, the results are slightly shifted: Due to the increased number of local minima, all algorithms tend do have a higher ANEES.
Since the ANEES of SM was too low in the 2D case, it now has the \enquote{best} ANEES for 3D.
However, we have to mention that this is just the result of two undesirable effects that compensate each other -- multiple minima and singular Jacobians.
If any measure is taken to improve the convergence, like a better initialization or graduated non-con\-vexity \cite{Yang2019}, it will become worse again.
Therefore, we are still convinced that our approach is able to deliver the best credibility if unmodeled properties like multiple minima are taken into account.

By comparing the run time of all GMM-based algorithms, we can see that MM is the most rapidly converging one -- which is not surprising due to its perfect linearity.
Interestingly, DCS is not noticeably faster, even though the evaluation of multiple GMM components comes with a bigger overhead. 
The proposed MSM is just slightly slower across all experiments, with a comparable number of iterations.
SM however, is far behind and converges between \num{2} and \num{10} times slower.
Its nonlinearity forces the optimizer to iterate often with many small or even unsuccessful steps.

In summary, we can say that the proposed Max-Sum-Mixture algorithm is the most consistent one in our evaluation.
It shows good convergence properties in the single cost function tests and the best accuracy combined with a good credibility in the point set registration experiment.

%% file: inc/conclusion.tex

\section{Conclusion}

In this work, we proposed a novel approach to implement Gaussian mixture models for robust least squares optimization.
These models provide the flexibility to describe almost arbitrary distributions in state estimation or to represent the ambiguity of assignment problems.
We demonstrated how to derive a well-posed loss function that represents the log-likelihood of a GMM mostly linear and without approximations.

By comparing against state-of-the-art methods, we showed the improved convergence properties of our Max-Sum-Mixture approach.
It was able to converge fast and accurate to the global minimum for a broad range of models, as our extensive Monte Carlo evaluation confirmed.
With point set registration, a practical application supported this results and showed also the superior combination of accuracy, credibility and computational efficiency that our approach provides.

We think that many state estimation problems in robotics can benefit from a well-converging Gaussian mixture formulation.
Therefore, we want to emphasize again the available open source implementation of the proposed algorithm. 

%% file: inc/appendix.tex

\appendices
\section{Proof of the Normalization Term} \label{app:proof}
In this section, we prove that our choice \autoref{eqn:msm_norm} of the normalization constant $\normMSM$ guarantees a negative log-likelihood above zero. 
\begin{IEEEproof}
	Define $\normMSM, \scaling_\iGMM \in \mathbb{R}^{\geq 0}$ and $\exponent_\iGMM \in \mathbb{R}^{\leq 0}$,\\
	where $\scaling_\iMax \exp(\exponent_\iMax) = \maxGMM \scaling_\iGMM \exp(\exponent_\iGMM)$ as defined in \autoref{eqn:msm_exp}\\
	and $\normMSM = \iGMMEnd \cdot \maxGMM \scaling_\iGMM$ as defined in \autoref{eqn:msm_norm}.
	\begin{IEEEeqnarray}{rCl}
		\scaling_\iMax \exp(\exponent_\iMax) & = & \maxGMM \scaling_\iGMM \exp(\exponent_\iGMM)                                                                                          \IEEEnonumber\\
		\scaling_\iMax \exp(\exponent_\iMax) & \geq & \frac{\sumGMM \scaling_\iGMM}{\iGMMEnd} \cdot \exp(\exponent_\iGMM)                                                                            \IEEEnonumber\\
		1                                          & \geq & \frac{\sumGMM \scaling_\iGMM \exp(\exponent_\iGMM)}{\iGMMEnd \cdot \scaling_\iMax \exp(\exponent_\iMax)}    
		\IEEEnonumber\\
		1                                          & \geq & \frac{\sumGMM \scaling_\iGMM  \exp(\exponent_\iGMM)}{\iGMMEnd \cdot \maxGMM (\scaling) \exp(\exponent_\iMax)}                       \IEEEnonumber\\
		\log(1)                                    & \geq & \log \left( \frac{\sumGMM \scaling_\iGMM  \exp(\exponent_\iGMM)}{\iGMMEnd \cdot \maxGMM (\scaling) \exp(\exponent_\iMax)} \right)   \IEEEnonumber\\
		0                                          & \leq & - \log \left( \frac{\sumGMM \scaling_\iGMM  \exp(\exponent_\iGMM)}{\iGMMEnd \cdot \maxGMM (\scaling) \exp(\exponent_\iMax)} \right) \IEEEnonumber\\
		0                                          & \leq & - \log \left( \sumGMM \frac{\scaling_\iGMM \exp (\exponent_\iGMM)}{\normMSM \exp(\exponent_\iMax)} \right)\IEEEnonumber \qquad \IEEEQEDhere
	\end{IEEEeqnarray}
\end{IEEEproof}

\section{Jacobians} \label{app:jacobian}

Here, we provide the Jacobian matrices for all cost functions that are given in \autoref{tab:cost}.
\\
For \emph{Max-Mixture} \cite{Olson2013} the Jacobian is
\begin{equation*}
\pfrac{\lossMM}{\residual} = 
\begin{bmatrix}[1.5]
\sqrtinfo_\iMax\\
\zero_{1,\iResEnd}
\end{bmatrix},
\end{equation*}
with $\zero_{1,\iResEnd}$ as the $[1 \times \iResEnd]$ zero matrix.
\\
For \emph{Sum-Mixture} \cite{Rosen2013} the Jacobian is
\begin{equation*}
\pfrac{\lossSM}{\residual} =  \frac{\sumGMM \left( \hat{\scaling}_\iGMM \exp (\exponent_\iGMM) \cdot \info_\iGMM \left( \residual -\mean_\iGMM \right) \right)}
{\sumGMM \hat{\scaling}_\iGMM \exp (\exponent_\iGMM) \cdot \sqrt{-2\lse(\vect{\hat{\scaling}}, \vect{\exponent})}},
\end{equation*}
with $\vect{\hat{\scaling}} = \left\lbrace \hat{\scaling}_1 \dots \hat{\scaling}_\iGMMEnd \right\rbrace $ and $\hat{\scaling}_\iGMM = \frac{\scaling_\iGMM}{\normSM}$.
\\
The proposed \emph{Max-Sum-Mixture} uses the modified scalings $\vect{\tilde{\scaling}} = \left\lbrace \tilde{\scaling}_1 \dots \tilde{\scaling}_\iGMMEnd \right\rbrace $
and exponents $\vect{\tilde{\exponent}} = \left\lbrace \tilde{\exponent}_1 \dots \tilde{\exponent}_\iGMMEnd \right\rbrace $, which are defined in detail as $\tilde{\scaling}_\iGMM = \frac{\scaling_\iGMM}{\normMSM}$ and $\tilde{\exponent}_\iGMM = \exponent_\iGMM - \exponent_\iMax$.
The Jacobian is
{\small\begin{equation*}
\pfrac{\lossMSM}{\residual} = 
\begin{bmatrix}[2]
\sqrtinfo_\iMax\\ \textstyle
\dfrac{\sumGMM \left( \tilde{\scaling}_\iGMM \exp (\tilde{\exponent}_\iGMM) \cdot \left(  \info_\iGMM \left( \residual -\mean_\iGMM \right)  - \info_\iMax \left( \residual -\mean_\iMax \right) \right) \right)  }
{\sumGMM \tilde{\scaling}_\iGMM \exp (\tilde{\exponent}_\iGMM) \cdot \sqrt{-2\lse(\vect{\tilde{\scaling}}, \vect{\tilde{\exponent}})}}
\end{bmatrix}.
\end{equation*}}%

%% file: root.bbl
\begin{thebibliography}{10}
\providecommand{\url}[1]{#1}
\csname url@samestyle\endcsname
\providecommand{\newblock}{\relax}
\providecommand{\bibinfo}[2]{#2}
\providecommand{\BIBentrySTDinterwordspacing}{\spaceskip=0pt\relax}
\providecommand{\BIBentryALTinterwordstretchfactor}{4}
\providecommand{\BIBentryALTinterwordspacing}{\spaceskip=\fontdimen2\font plus
\BIBentryALTinterwordstretchfactor\fontdimen3\font minus
  \fontdimen4\font\relax}
\providecommand{\BIBforeignlanguage}[2]{{%
\expandafter\ifx\csname l@#1\endcsname\relax
\typeout{** WARNING: IEEEtran.bst: No hyphenation pattern has been}%
\typeout{** loaded for the language `#1'. Using the pattern for}%
\typeout{** the default language instead.}%
\else
\language=\csname l@#1\endcsname
\fi
#2}}
\providecommand{\BIBdecl}{\relax}
\BIBdecl

\bibitem{Olson2013}
E.~Olson and P.~Agarwal, ``Inference on networks of mixtures for robust robot
  mapping,'' \emph{Int. Journal of Robotics Research}, vol.~32, no.~7, pp.
  826--840, 2013.

\bibitem{Rosen2013}
D.~M. Rosen, M.~Kaess, and J.~J. Leonard, ``Robust incremental online inference
  over sparse factor graphs: Beyond the {Gauss}ian case,'' in \emph{Proc. of
  Int. Conf. on Robotics and Automation (ICRA)}, 2013, pp. 1025--1032.

\bibitem{Agarwal}
S.~Agarwal, K.~Mierle, and Others, ``{Ceres Solver},''
  \url{http://ceres-solver.org}.

\bibitem{Dellaert}
F.~Dellaert and Others, ``{GTSAM},''
  \url{http://research.cc.gatech.edu/borg/gtsam}.

\bibitem{Huber1964}
P.~J. Huber, ``Robust estimation of a location parameter,'' \emph{The Annals of
  Mathematical Statistics}, vol.~35, pp. 73--101, 1964.

\bibitem{Agarwal2013}
P.~Agarwal, G.~D. Tipaldi, L.~Spinello, C.~Stachniss, and W.~Burgard, ``Robust
  map optimization using dynamic covariance scaling,'' in \emph{Proc. of Int.
  Conf. on Robotics and Automation (ICRA)}, 2013, pp. 62--69.

\bibitem{Agamennoni2015}
G.~Agamennoni, P.~Furgale, and R.~Siegwart, ``Self-tuning m-estimators,'' in
  \emph{Proc. of Int. Conf. on Robotics and Automation (ICRA)}, 2015, pp.
  4628--4635.

\bibitem{Chebrolu2021}
N.~Chebrolu, T.~L{\"a}be, O.~Vysotska, J.~Behley, and C.~Stachniss, ``Adaptive
  robust kernels for non-linear least squares problems,'' \emph{Robotics and
  Automation Letters (RA-L)}, 2021.

\bibitem{Suenderhauf2012a}
N.~S{\"u}nderhauf and P.~Protzel, ``Switchable constraints for robust pose
  graph slam,'' in \emph{Proc. of Int. Conf. on Intelligent Robots and Systems
  (IROS)}, 2012, pp. 1879--1884.

\bibitem{Pfingsthorn2012}
M.~Pfingsthorn and A.~Birk, ``Simultaneous localization and mapping with
  multimodal probability distributions,'' \emph{The Int. Journal of Robotics
  Research}, vol.~32, no.~2, pp. 143--171, 2012.

\bibitem{Pfeifer2016}
T.~Pfeifer, P.~Weissig, S.~Lange, and P.~Protzel, ``Robust factor graph
  optimization -- a comparison for sensor fusion applications,'' in \emph{Proc.
  of Int. Conf. on Emerging Technologies and Factory Automation (ETFA)}, 2016,
  pp. 1--4.

\bibitem{Stoyanov12a}
T.~Stoyanov, M.~Magnusson, H.~Andreasson, and A.~J. Lilienthal, ``Fast and
  accurate scan registration through minimization of the distance between
  compact {{3D NDT}} representations,'' \emph{The Int. Journal of Robotics
  Research}, vol.~31, no.~12, pp. 1377--1393, 2012.

\bibitem{Hsiao2019}
M.~Hsiao and M.~Kaess, ``Mh-isam2: Multi-hypothesis isam using bayes tree and
  hypo-tree,'' in \emph{Proc. of Int. Conf. on Robotics and Automation (ICRA)},
  Montreal, QC, Canada, 2019, pp. 1274--1280.

\bibitem{Fourie2016}
D.~Fourie, J.~Leonard, and M.~Kaess, ``A nonparametric belief solution to the
  bayes tree,'' in \emph{Proc. of Int. Conf. on Intelligent Robots and Systems
  (IROS)}, 2016, pp. 2189--2196.

\bibitem{Madsen2004}
K.~Madsen, H.~Nielsen, and O.~Tingleff, ``Methods for nonlinear least squares
  problems,'' Technical University of Denmark, Tech. Rep., 2004.

\bibitem{Pfeifer}
T.~Pfeifer and Others, ``{libRSF},''
  \url{https://github.com/TUC-ProAut/libRSF}.

\bibitem{Barjenbruch15}
M.~Barjenbruch, D.~Kellner, J.~Klappstein, J.~Dickmann, and K.~Dietmayer,
  ``Joint spatial- and {{Doppler}}-based ego-motion estimation for automotive
  radars,'' in \emph{Proc. of {{Int}}. {{Symposium}} on {{Intelligent
  Vehicles}} ({{IV}})}, 2015, pp. 839--844.

\bibitem{Jian11}
B.~Jian and B.~C. Vemuri, ``Robust {{Point Set Registration Using Gaussian
  Mixture Models}},'' \emph{IEEE Trans. on Pattern Analysis and Machine
  Intelligence}, vol.~33, no.~8, pp. 1633--1645, 2011.

\bibitem{Carlone14}
L.~Carlone and A.~Censi, ``From {{Angular Manifolds}} to the {{Integer
  Lattice}}: {{Guaranteed Orientation Estimation With Application}} to {{Pose
  Graph Optimization}},'' \emph{IEEE Transactions on Robotics}, vol.~30, no.~2,
  pp. 475--492, 2014.

\bibitem{Shalom01}
Y.~Bar-Shalom, X.~R. Li, and T.~Kirubarajan, \emph{Estimation with
  {{Applications}} to {{Tracking}} and {{Navigation}}: {{Theory Algorithms}}
  and {{Software}}}, 1st~ed.\hskip 1em plus 0.5em minus 0.4em\relax {John Wiley
  \& Sons, Inc.}, 2001.

\bibitem{li02}
X.~R. Li, Z.~Zhao, and V.~P. Jilkov, ``Estimator’s credibility and its
  measures,'' in \emph{Proc. of {{Int}}. {{Federation}} of {{Automatic
  Control}} ({{IFAC}})}, 2002.

\bibitem{Yang2019}
H.~Yang, P.~Antonante, V.~Tzoumas, and L.~Carlone, ``Graduated non-convexity
  for robust spatial perception: From non-minimal solvers to global outlier
  rejection,'' \emph{Robotics and Automation Letters (RA-L)}, pp. 1127--1134,
  2019.

\end{thebibliography}
